\documentclass{article}
\usepackage{spconf,amsmath,graphicx}
\usepackage{graphicx}
\usepackage{multirow}
\usepackage{catchfile}
\usepackage{longtable}
\usepackage{caption}
\usepackage{subcaption}
\graphicspath{{./figure/}}
\usepackage[dvipsnames]{xcolor}

\title{DECIDING WHETHER TO ASK CLARIFYING QUESTIONS IN LARGE-SCALE SPOKEN LANGUAGE UNDERSTANDING}
\name{Joo-Kyung Kim, Guoyin Wang, Sungjin Lee, Young-Bum Kim}
\address{Amazon Alexa AI}
\begin{document}
%\ninept
%
\maketitle
\begin{abstract}
A large-scale conversational agent can suffer from understanding user utterances with various ambiguities such as ASR ambiguity, intent ambiguity, and hypothesis ambiguity.
When ambiguities are detected, the agent should engage in a clarifying dialog to resolve the ambiguities before committing to actions.
However, asking clarifying questions for all the ambiguity occurrences could lead to asking too many questions, essentially hampering the user experience.
To trigger clarifying questions only when necessary for the user satisfaction, we propose a neural self-attentive model that leverages the hypotheses with ambiguities and contextual signals.
We conduct extensive experiments on five common ambiguity types using real data from a large-scale commercial conversational agent and demonstrate significant improvement over a set of baseline approaches.
\end{abstract}
\begin{keywords}
spoken language understanding, clarifying questions, user satisfaction, self-attention
\end{keywords}

\section{Introduction}
\label{sec:intro}

The spoken language understanding (SLU) task of a large-scale conversational agent goes through a set of components such as automatic speech recognition (ASR), natural language understanding (NLU) including domain classification, intent classification, and slot-filling \cite{Tur2011}, and skill routing (SR) \cite{YBKim2018b,JKKim2020}, which is responsible for selecting an appropriate service provider to handle the user request.%and picking one of NLU results to feed it with. 

When a user interacts with the agent, the underlying systems may not be able to understand what the user actually wants if the utterance is ambiguous.
Ambiguity comes from ASR when audio cannot be recognized correctly (\emph{e.g.}, audio quality issues can cause ASR to confuse ``Five minute timer'' and ``Find minute timer'');
it comes from NLU when the user's request cannot be interpreted correctly (\emph{e.g.}, ``Garage door'' could mean open it or close it);
it comes from SR when it is not possible to confidently select the best experience between multiple valid service providers (\emph{e.g.}, ``Play frozen'' can mean playing \texttt{video}, \texttt{soundtrack}, or \texttt{game}), and so on.
Ignoring such ambiguities from upstream components can pass incorrect signals to the downstream and lead to an unsatisfactory user experience.

\begin{figure}[t]
    \centering
    \begin{subfigure}[b]{1\columnwidth}
        \centering
        \resizebox{\columnwidth}{!}{
        \begin{tabular}{lll}
        \hline
        \textbf{ASR} & \textbf{Recognized Dialog} & \textbf{Label} \\
        \hline
        & \textbf{U:} Set a timer for \textcolor{blue}{15} minutes \\
        Correct & \textbf{S:} 15 minutes, right? & \textcolor{blue}{\textbf{Unnecessary}}\\
        & \textbf{U:} Yes. \\
        & \textbf{S:} 15 minutes, starting now. \\
        \hline
        & \textbf{U:} Set a timer for \textcolor{red}{50} minutes \\
        Incorrect & \textbf{S:} 50 minutes, right? & \textcolor{red}{\textbf{Necessary}}\\
        & \textbf{U:} No, 15 minutes \\
        & \textbf{S:} 15 minutes, starting now. \\
        \hline
        \end{tabular}
        }
    \end{subfigure}
    \caption{Examples of \textcolor{blue}{Unnecessary} and \textcolor{red}{Necessary} clarifying turns when a user actually said ``Set a timer for \textbf{15} minutes'' but ASR ambiguity exists between 15 and 50, where the most confident ASR is (top: correct (\textcolor{blue}{15}), bottom: incorrect (\textcolor{red}{50}).)}
    \label{fig:example}
\end{figure}

Thus, when the agent is unsure due to ambiguity, it should engage in a clarifying dialog before taking actions.
However, asking clarifying questions for all the detected ambiguous utterances can end up spamming users with too many redundant questions, resulting in poor user experiences as shown in Fig.~\ref{fig:example}.

In SLU systems, the ambiguities are typically detected by applying thresholds to the system components' confidence scores along with certain heuristics \cite{Komatani2000,Gabsdil2003,Skantze2005,Bohus2005,Ayan2013,Stoyanchev2015,Zukerman2015}.
However, using thresholds to properly filter ambiguities is a practically difficult task in large-scale SLU.
First, it is non-trivial for humans to define fine-grained thresholds for the upstream components or rules for deciding clarification considering dialog and ambiguity context information. It is more complicated to make the decisions when multiple ambiguities from different components co-occur since each ambiguity is entangled to the other ambiguities in many cases. Therefore, relying on the thresholding based approach is not scalable in recent general-purposed conversational agents such as Amazon Alexa, Google Assistant, and Apple Siri since a huge number of various dialog contexts and ambiguity situations should be considered.
Second, each system component is differently supervised and the confidence thresholding is separately decided in large-scale pipelined SLU systems \cite{Takanobu2020}. For example, an ASR prediction with softmax outputs can be considered ambiguous when the second best ASR confidence is above 30\% while an intent classification with sigmoid outputs is defined to be ambiguous when the second best intent's confidence is above 80\%.
In addition, each upstream component can be independently updated in large-scale systems while previous studies such as \cite{Ayan2013} and \cite{Zukerman2015} assume that each component is fixed or the components can be holistically manageable.
Therefore, many of the detected ambiguities are indeed false positives (\emph{i.e.}, unambiguous) with a different ratio for each SLU component. 
Lastly, even if obvious ambiguities exist, clarifying dialogs are unnecessary if the top prediction is correct in terms of user satisfaction. It was shown that more than 60\% of ASR errors do not need clarification since many of the errors do not influence the end-to-end performance in a speech-to-speech translation system \cite{Stoyanchev2012} and we also observe a similar tendency from our work in Section \ref{sec:experiments}.

There are various studies about how to compose clarifying questions\footnote{e.g., asking either reprise questions for targeted clarification or generic questions such as repeat or rephrase requests dependent on the occurred ambiguity contexts.} and their effectiveness in SLU systems when ambiguities exist \cite{Gabsdil2003,liu2014,Stoyanchev2014,Kiesel2018}.
Also, clarifying questions on other tasks such as Q\&A \cite{Rao2018,Xu2019,Kumar2020,Min2020} and information retrieval \cite{Aliannejadi2019,Zamani2020,Padmakumar2021} are being actively studied.
However, none of them specifically focus on initiating clarifying interactions only when necessary in the interest of preventing user experience degradation, which is crucial in large-scale SLU systems.

To address the issue of deciding whether to ask clarifying questions in large-scale SLU systems, we propose a unified neural self-attentive model that makes a global decision on whether to trigger a clarifying question considering ambiguity occurrence information and various contextual signals. We show that the self-attentive representations of the top hypothesis and the aggregated alternative hypotheses from a hypothesis reranker \cite{Robichaud2014,Khan2015,YBKim2018b,JKKim2020} are effective for dealing with the ambiguities from SLU.

Given the fact that a large-scale conversational system supports various devices, languages, and application components, providing access to a wide variety of skills \cite{Kumar2017,YBKim2018a,JKKim2018}, it is not scalable to rely on manually annotated data to train and evaluate the model. Instead, we leverage a user satisfaction model, which has recently attracted significant attention in both academia and industry \cite{Hashimoto2018,Hancock2019,Park2020}, to generate ground-truth labels at scale. The user satisfaction model we use \cite{Park2020} marks defective turns by examining the input utterance, the system response, and the user's implicit/explicit feedback of their following turns. Having turn-level defect labels, our model is supervised to not trigger a clarifying question when the agent is likely to deliver satisfying experience even if ambiguities are detected from the upstream SLU components.

In this paper, we define five ambiguity types that are popular in the SLU task (see Section~\ref{ssec:ambiguity_type}), conduct extensive experiments using real data from a large-scale commercial conversational system, and demonstrate significant improvements over several baseline approaches in reducing unnecessary clarifying interactions.

\section{Ambiguities in SLU}
\label{sec:ambiguity}
Our task is to determine whether to ask clarifying questions when ambiguity signals are captured by any SLU components in the user utterance.
In this section, we describe how an input utterance is interpreted with different hypotheses in SLU, the ambiguity types we are dealing with, and how ground-truths of the ambiguous utterances are assigned in our work.

\subsection{Hypothesis representations}
\label{ssec:hyp_rep}
In large-scale SLU, it is common to represent various possible interpretations of an input utterance as \textit{hypotheses}, each of which contains the outputs and the confidence scores of upstream components such as ASR, domain, intent, and slot-filling results \cite{Robichaud2014,Khan2015,YBKim2018b,JKKim2020}.
Given the hypothesis list as the input, a hypothesis ranker (HypRank) is used to rank the input hypotheses.
Table \ref{tab:hypotheses} shows an example of the hypotheses from HypRank.
\begin{table}[t]
\small
\resizebox{0.48\textwidth}{!}{%
\begin{tabular}{|l|l|l|l|l|l|l|}
\hline
\# &
  ASR &
  \begin{tabular}[c]{@{}l@{}}ASR\\ Conf\end{tabular} &
  Intent &
  \begin{tabular}[c]{@{}l@{}}Intent\\ Conf\end{tabular} &
  Slots &
  \begin{tabular}[c]{@{}l@{}}Hyp\\ Conf\end{tabular} \\ \hline\hline
1 & harry potter & 0.9 & PlayVideo  & 0.95 & videotitle & 0.9 \\ \hline
2 & harry potter & 0.9 & ReadBook   & 0.75 & booktitle  & 0.8 \\ \hline
3 & harry potter & 0.9 & SoundTrack & 0.94 & albumtitle & 0.6 \\ \hline
4 & harry potter & 0.9 & GetWiki    & 0.7  & entity     & 0.4 \\ \hline
5 & larry potter & 0.6 & ReadBook   & 0.65 & booktitle  & 0.1 \\ \hline
\end{tabular}%
}
\caption{An example of the ranked hypothesis list from HypRank given \emph{harry potter} as the utterance.}
\label{tab:hypotheses}
\end{table}

Then, given the HypRank output, which is a ranked hypothesis list, the top ranked hypothesis is chosen as the final SLU decision for unambiguous utterances.
For ambiguous utterances, since it is unconfident whether the top hypothesis would be promising, clarification interactions allow choosing a non-top/alternative hypothesis.
In this work, we focus on deciding whether clarification is necessary or not for the ambiguous utterances.

\subsection{Ambiguity Types}
We first define five common types of ambiguities in SLU as follows:
\begin{table*}[]
\small
\begin{tabular}{|l|l|l|l|l|}
\hline
Ambiguity & User utterance             & Potential clarifying question                            & User response                    & System response or \textless{}action\textgreater{} \\ \hline\hline
ASR            & Set a thirty minute timer  & Do you mean thirteen or thirty?                & Thirteen                         & Start a thirteen minute timer                      \\ \hline
IC             & Turn on off                & Do you mean turn on or turn off?               & Off                              & \textless{}Turn off the agent\textgreater{}        \\ \hline
HC             & Get me a ride              & Do you want Uber or Lyft?                      & Uber                             & Finding a uber driver                              \\ \hline
SNR            & Turn on the ligXXX (noisy) & Sorry, could you repeat it? & Turn on the light (clear) & \textless{}Turn on the light\textgreater{}         \\ \hline
TRUNC          & Turn on the                & Sorry, turn on what?       & The fan                              & \textless{}Turn on the fan\textgreater{}           \\ \hline
\end{tabular}
\caption{Clarifying dialog examples}
\label{tab:dialog_examples}
\end{table*}

\label{ssec:ambiguity_type}
\begin{itemize}
\item \textbf{ASR}: Two ASR outcomes are regarded as ambiguous when their edit distance is 1, their ASR confidences are close, and they produce different slot values. \emph{e.g.}, ``thirteen minutes'' vs ``thirty minutes''.

\item \textbf{Similar Intent Confidences} (IC): The intent of an utterance is ambiguous. \emph{e.g.}, ``turn on off'' is ambiguous since both \texttt{TurnOnIntent} and \texttt{TurnOffIntent} can have high confidences by the intent classifier.

\item \textbf{Similar Hypothesis Confidences} (HC): The final hypothesis confidences from HypRank \cite{YBKim2018b,JKKim2020} are similar. \emph{e.g.}, ``get me a ride'' can have similar confidences for the hypotheses associated with \texttt{UBER} and \texttt{LYFT} as the service providers.

\item \textbf{Signal to Noise Ratio} (SNR): When the acoustic noise level is very high, it is not clear whether we can trust the ASR output even if the ASR confidence is sufficiently high.

\item \textbf{Utterance Truncation} (TRUNC): An utterance can be recognized missing the later tokens due to slow speaking or ASR errors. \emph{e.g.}, if a user said ``Music composed by Mozart'' but only ``Music composed by'' are recognized, the missed token should be clarified. In this work, we regard utterances ending with articles (``a'', ``an'', ``the''), some possessive pronouns (\emph{e.g.}, ``my''), or some prepositions (\emph{e.g.}, ``by'') as truncated.
\end{itemize}

Table \ref{tab:dialog_examples} shows the clarifying dialog examples for different ambiguity types, which demonstrates how clarifying questions can help resolve the ambiguities.

\subsection{Ground-truth Labeling}
\label{ssec:gt_labeling}
It is difficult to decide whether a clarifying question would be helpful or not when ambiguities exist since each ambiguity is with a different occurrence condition, multiple ambiguities can co-occur, and the top predictions are correct in many cases even if ambiguities exist.
In this work, we regard ambiguous utterances with unsatisfactory results as those need clarifications, and vice versa. The rationale is that if a user is unsatisfied with the top predicted hypothesis from the HypRank when ambiguities exist, the user could have been satisfied by allowing the user to choose another hypothesis.

We use the log data from a conversational agent system, where each utterance is assigned to be either satisfactory or unsatisfactory by a user satisfaction metric \cite{Park2020,Hashimoto2018,Hancock2019}. Specifically, we use a model-based metric described in \cite{Park2020}, which utilizes the current turn's utterance and the response as well as the follow-up turn's utterance to judge whether the current turn is satisfactory or not.

Our labeling method is a weak supervision approach assuming no clarifying questions exist in the log data. If the log data already include turns with clarifying questions, we can identify whether the questions were helpful or not. For example, if a user selected the top hypothesis in the clarification, it was unnecessary to ask since the top hypothesis could be chosen without the clarification. Oppositely, if the user chose the other hypothesis, then the clarification was useful evading unwanted top hypothesis. Formally speaking, the ground-truths can be set with counterfactual learning using the logged data, but this is beyond the scope of this work.

\begin{figure*}[t]
	\centering
	\includegraphics[width=0.6\textwidth]{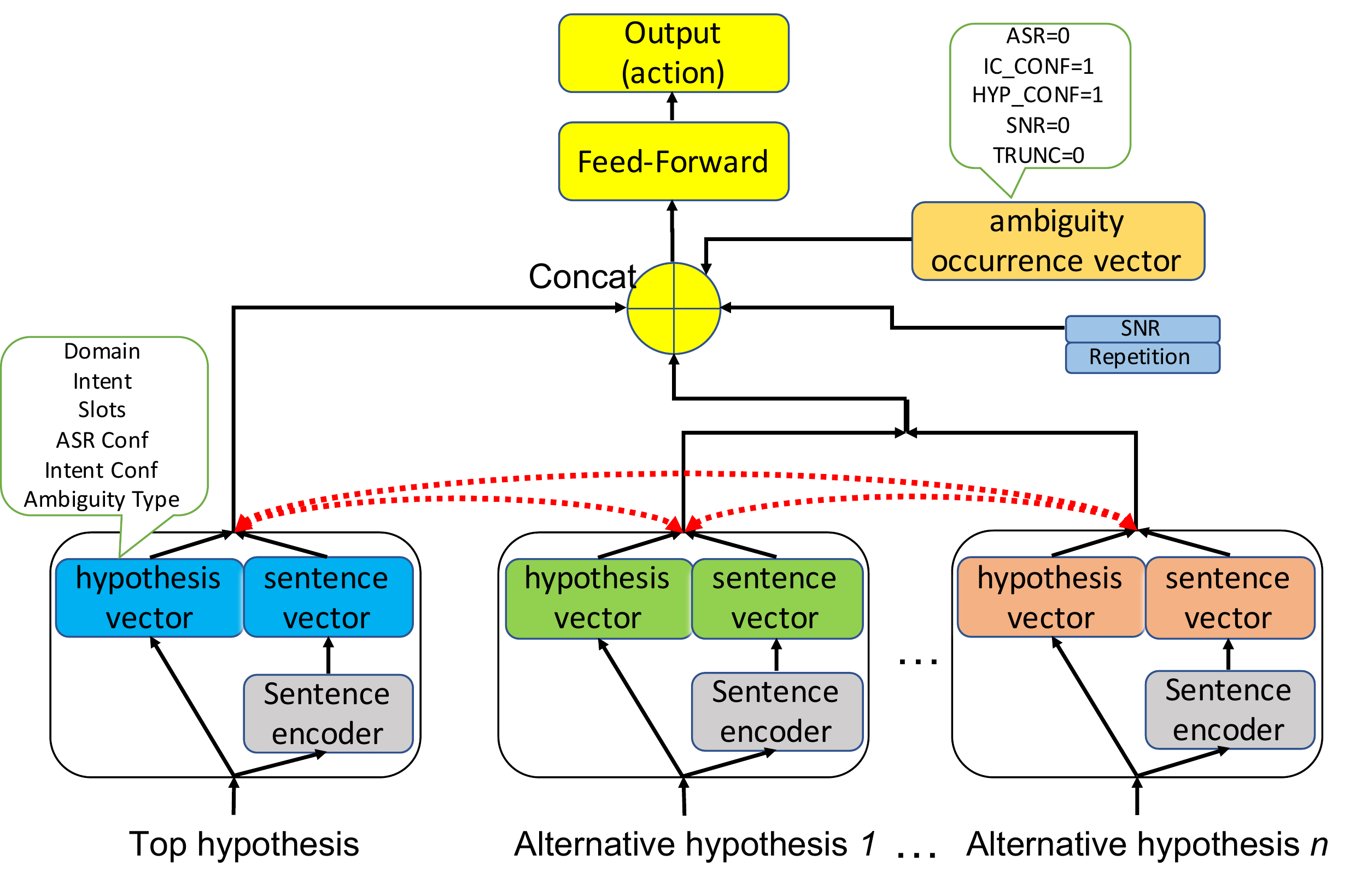}
	\caption{\small The model architecture. The top hypothesis and the alternative hypotheses corresponding to the occurred ambiguities are extracted from the HypRank output list and used as the input sequence. On top of the vector sequence, self-attention mechanism is used to produce contextualized hypotheses representations. Then the top hypothesis output, the summation of the alternative hypothesis outputs, the ambiguity occurrence vector, SNR value, and repetition value are concatenated to represent the entire information and it is transformed to an output vector for deciding `ask' or `not'.}
	\label{fig:arch}
\vspace{-1mm}
\end{figure*}

\section{Model}
Figure \ref{fig:arch} shows the overall architecture of the proposed model deciding whether to ask clarifying questions or not.

The input to our proposed model is a subset of the HypRank hypotheses described in Section \ref{ssec:hyp_rep}. The top predicted hypothesis is always included in our model input since it is what to be compared with the alternative hypotheses through clarification dialogs.
Then, for each occurred ambiguity, we add the most confident alternative hypothesis corresponding to the ambiguity.
For the example in Table \ref{tab:hypotheses}, assuming 0.8 is the threshold to represent ASR, IC, and HC ambiguity occurrences, and SNR and TRUNC ambiguities did not occur, hypothesis \#2 is with the highest confidence among the hypotheses corresponding to HC ambiguity and \#3 is the one corresponding to IC ambiguity. Since the hypothesis with different transcript has lower ASR confidence (0.6) than the threshold (0.8), we do not use it as an alternative hypothesis. Therefore, the input sequence to the proposed model consists of hypothesis \#1 as the top hypothesis, and \#2 and \#3 as the alternative hypotheses.
\footnote{SNR and Trunc ambiguities do not have corresponding alternative hypotheses since those ambiguities could be resolved by generating new hypotheses based on additional information from clarification interactions. Therefore, we represent the corresponding hypothesis with \textit{$\left<unk\right>$} vector for each hypothesis elements.}

As the model input, each hypothesis is represented as a concatenated vector of ASR, ASR confidence, intent confidence, domain, intent, slots, and ambiguity type.\footnote{We do not include the hypothesis confidence as an input feature since there exist utterances without the scores due to rule-based or shortlister only hypothesis decision.} ASR is represented by the output summation of a single layered standard multi-head transformer encoder \cite{Vaswani2017}\footnote{We empirically find 4 attention heads shows the best performance.}, where the word embedding is initialized with GloVe \cite{Pennington2014}. ASR confidence is a scalar value normalized to be between 0 and 1. A vector for slots is represented as a sum of matched slot key vectors similarly to \cite{YBKim2018b}. Domain, intent, and ambiguity type\footnote{The top hypothesis's ambiguity type is denoted as \texttt{TOP} to differentiate it with the alternative hypotheses' ambiguity types such as \texttt{ASR}, \texttt{IC}, and \texttt{HC}.} are also vectorized with embeddings.\footnote{The effectiveness of these features is shown in Section \ref{ssec:ablation}.}
On top of the hypothesis vector sequence, we obtain a contextualized vector sequence using self-attention with a transformer encoder. For this self-attention, inspired by Set Transformer \cite{Lee2019}, we do not utilize position encoding since the order of alternative hypotheses for different ambiguities is not informative for the model's decision.

From the contextualized hypotheses, we obtain the top hypothesis's representation and the sum of the alternative hypothesis representation to be used as the inputs to the final prediction layer.
Summation of the alternatives is necessary since the number of the alternative hypotheses (\emph{i.e.} \# occurred ambiguities) varies for each utterance and they should be aggregated to be used as an input representation.
While the majority of self-attentive models for other tasks use single representation aggregated over all the elements in the given sequence, we observe that separating the top hypothesis representation and the aggregated representation over the alternative hypotheses performs better due to different aspects of the top hypothesis and the alternative hypotheses in terms of deciding clarification or not.

In addition to the hypothesis representations, we also use other signals: SNR, which is a scalar value normalized to be between -1 and 1, ambiguity occurrence vector, which is a concatenation of binary values representing the occurred ambiguities, and a binary signal representing repetition of the previous user utterance, which is a common indicator that the same utterance was wrongly recognized or unsatisfactory previously.
All these vectors are concatenated and transformed to an output vector through a feed-forward network.

\section{Experiments}
\label{sec:experiments}

\subsection{Datasets}
% \usepackage{multirow}

\begin{table*}[ht!]
\small
\centering
\begin{tabular}{|c|c|c|c|c|c|c|c|c|c|}
\hline
\multirow{2}{*}{Ambiguity type} & \multicolumn{3}{c|}{Train} & \multicolumn{3}{c|}{Valid} & \multicolumn{3}{c|}{Test} \\ \cline{2-10} 
                                & Total & Ask  & No ask  & Total   & Ask    & No ask  & Total   & Ask   & No ask  \\ \hline \hline 
ASR                             & 4.3M  & 763K & 4.2M   & 477K & 82K  & 395K     & 4.6M  & 780K   & 3.8M   \\ \hline
IC                              & 4.2M  & 612K & 3.6M   & 464K & 67K  & 397K     & 4.5M  & 491K   & 4M    \\ \hline
HC                              & 265K  & 77K  & 188K   & 29K  & 8K   & 21K      & 343K  & 92K   & 251K   \\ \hline
SNR                             & 16.6M & 3.4M & 13.2M  & 1.8M & 375K & 1.4M     & 18.2M & 3.8M   & 14.4M  \\ \hline
TRUNC                           & 2.6M  & 1.6M & 1M     & 292K & 178K & 114K     & 2.8M  & 1.7M   & 1.1M     \\ \hline
Total                           & 28.6M & 6.5M & 22.2M  & 3M   & 710K & 2.3M     & 30.4M & 6.9M   & 23.6M  \\ \hline
\end{tabular}
\caption{Ambiguity dataset sizes.}
\label{tab:data}
\vspace{-1mm}
\end{table*}
To the best of our knowledge, there is no existing public dataset for asking clarification questions including ASR related features such as ASR confidences and SNR values.
Based on an assessment of a randomly sampled ambiguous utterances from a conversational AI system, we estimate that about 23\% ambiguous traffic should be resolved for user satisfaction through a clarification dialog.
To show statistical significance on the evaluation results and to make the data split similar to real deployment scenarios, we construct a large test set by selecting the second half of the data based on time stamp. We then randomly split the first half of data to training/validation sets with 9:1 ratios. The detailed statistics for each ambiguity type are summarized in Table \ref{tab:data}. For example, there are total 4.6M utterances with ASR ambiguity in the test set. However, only 780K of them are worth to clarify for better user satisfaction and the remaining 3.8M do not need to clarify since these are satisfactory to the users even though they are ambiguous.
The ratio of `ask' labels varies for different ambiguity types due to different criteria and thresholds in ambiguity detection.

\subsection{Experiment Setup}
\label{ssec:setup}
\begin{figure}
    \small
     \centering
    %  \begin{subfigure}[]{0.15\textwidth}
    %      \centering
    %      \includegraphics[width=\textwidth]{alt_att.png}
    %      \caption{}
    %      \label{fig:alt_att}
    %  \end{subfigure}
    %  \hfill
     \begin{subfigure}[]{0.1\textwidth}
         \centering
         \includegraphics[height=0.8in]{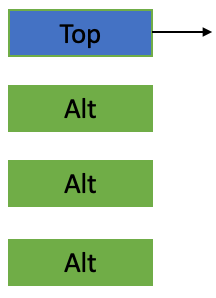}
         \caption{\emph{no-alt}}
         \label{fig:no_alt}
     \end{subfigure}
     \hfill
     \begin{subfigure}[]{0.1\textwidth}
         \centering
         \includegraphics[height=0.8in]{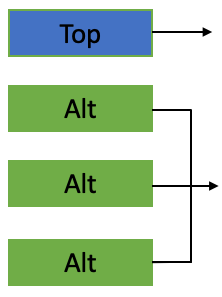}
         \caption{\emph{alt-avg}}
         \label{fig:alt_avg}
     \end{subfigure}
     \hfill
     \begin{subfigure}[]{0.1\textwidth}
         \centering
         \includegraphics[height=0.8in]{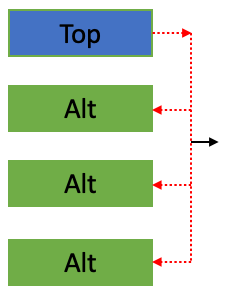}
         \caption{\emph{crs-att}}
         \label{fig:crs_att}
     \end{subfigure}
     \hfill
     \begin{subfigure}[]{0.1\textwidth}
         \centering
         \includegraphics[height=0.8in]{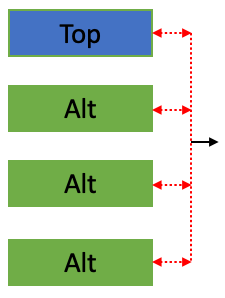}
         \caption{\emph{self-att}}
         \label{fig:all_att}
     \end{subfigure}
        \caption{Attention structure illustration. Solid arrows and dotted arrows denote using the outputs and using the attentions, respectively, and \texttt{Top} and \texttt{Alt} denote the top hypothesis and each alternative hypothesis, respectively:
         (a) using the top only (b) using the average alternatives (c) the top hypothesis as the key for the cross attention over the alternatives (d) self-attention over all the hypotheses.}
        \label{fig:att_fig}
\vspace{-3mm}
\end{figure}
In a high level, we consider three approaches in our experiments, 
(\emph{i}) asking questions for every ambiguity occurrence, denoted as \emph{Always}; 
(\emph{ii}) utilizing the top hypothesis and the context information as the input to the output layer, denoted as \emph{No-alt} in Figure~\ref{fig:no_alt};
(\emph{iii}) using all the top and the alternative hypotheses and the context information.

In the third category, we try different types of alternative hypothesis aggregation.
The simplest one is using the top hypothesis vector and simple average over the alternative hypothesis vectors, denoted as \emph{Alt-avg} in Figure~\ref{fig:alt_avg}.
To more effectively represent contextualized information of the hypotheses, we try two types of attention mechanisms:
(1) representing alternative hypotheses with cross attention given the top hypothesis as the key, denoted as \emph{Crs-att} in Figure~\ref{fig:crs_att}.
(2) a complete self-attention over all the hypotheses is \emph{Self-att} in Figure~\ref{fig:all_att}.

When using both the top hypothesis and the alternative hypothesis aggregation, the previous approaches concatenate the top and the alternative vectors to be used as the input to the final layer. We also try their summation, denoted as \emph{Self-sum} to check whether separately representing the top and the alternatives in the final layer is better or not.

\subsection{Implementation Details}
We train each model with ADAM optimizer~\cite{Kingma2015} for 20 epochs and select the best model based on performance on validation set. The dimensionality of hypothesis components such as domain, intent, and slot vectors and utterance vectors are set to 100. The other hyperparameters for self-attention and positional embedding are identical to the default values in \cite{Vaswani2017}.

\subsection{Experiment Results}
Note that precision can evaluate the model's ability to avoid unnecessary clarifications and recall can measure the ability to ask clarifications when necessary. Hence, we use F1 score, which balances the two metrics, to evaluate the model performance. To make a thorough evaluation, we evaluate both F-1 for each ambiguity type and the overall F-1 for all the types.

The relative F-1 scores over all aforementioned models are summarized in Table \ref{tab:result}.\footnote{ Due to internal confidentiality policy, we report relative F-1 scores, $\left(f_x-f_{always}\right)/f_{always}$, where $f_x$ and $f_{always}$ denote F-1 scores of model $x$ and model $Always$, respectively.}
Since \emph{Always} always asks clarifying questions for the ambiguous utterances, its F-1 score is the lowest due to low precision even though recall is 100\%. As aforementioned, about 23\% of all the ambiguous utterances need clarification in our experiment setting, thereby the F-1 score of all the ambiguities is around 37\%, where each ambiguity's F-1 is between 20\% to 70\%. Therefore, using any of the tested models shows significantly better performance.

Compared to \emph{No-alt} model, which does not utilize the alternative hypothesis information, \emph{Alt-avg} does not significantly improve the performance.
However, the other models using attention mechanisms to represent the alternative hypotheses significantly outperform \emph{No-alt} model. This indicates that properly represented alternative hypothesis information is an important factor in the model decision.
Also, using self-attention, \emph{Self-att}, is again significantly better than using cross-attention with the top hypothesis as the key, \emph{Crs-att}. This shows that the fully contextualized hypothesis representations are helpful.
We further deepen the model by considering 2-layer of self-attention, which further improves the performance.
Such good performance verifies the effectiveness of our self-attention based model in asking clarification question task.
In addition, relatively poor performance of \emph{Self-sum} reflects that the proposed architecture, which separately represents the top hypothesis and the aggregated alternative hypotheses with concatenation, is more proper for our task than singly aggregated representation, which is more common in self-attentive architectures. This empirically demonstrates the top hypothesis and the alternative hypotheses play different roles in the model decision, thereby separating them is more helpful.

% \usepackage{multirow}

\begin{table}[t]
\small
\begin{tabular}{|c|c|c|c|c|c|c|}
\hline
Model & All   & ASR  & IC    & HC    & SNR & TRUNC  \\ \hline \hline
Always     & 0 & 0 & 0 & 0 & 0 & 0\\ \hline
No-alt     & 77.79 & 71.63 & 99.88  & 201.86 & 9.28 & 88.76 \\ \hline
Alt-avg    & 77.81 & 72.43 & 98.88  & 212.28 & 8.90 & 74.89 \\ \hline
Crs-att    & 78.05 & 72.35 & 99.23  & 212.18 & 9.09 & 79.02 \\ \hline
Self-sum   & 78.21 & 70.67 & 101.49 & 214.31 & 9.30 & 89.43 \\ \hline
Self-att   & 79.61 & 73.70 & 103.03 & 214.01 & 9.42 & \textbf{90.42} \\ \hline
Self-att2  & \textbf{81.09} & \textbf{75.91} & \textbf{105.34} & \textbf{216.80} & \textbf{9.79} & 87.36 \\ \hline
\end{tabular}
\caption{Relative F-1 \% improvements of different model approaches compared to \emph{Always} scores. \emph{Self-att2} denotes two layers of self-attention.}
\label{tab:result}
\end{table}

\subsection{Ablation Study}
\label{ssec:ablation}
\begin{table}[t]
\small
\begin{tabular}{|c|c|c|c|c|c|c|}
\hline
Model          & All            & ASR           & IC             & HC             & SNR          & TRUNC \\ \hline \hline
No-hyp         & -37.74 & -49.81 & -32.90 & -28.84 & -31.11 & -10.08 \\ \hline
ASR            & -35.90 & -46.72 & -34.30 & -24.30 & -30.11 & -7.96  \\ \hline
No-sent        & -16.01 & -21.69 & -31.70 & -9.55  & -5.68  & -31.18 \\ \hline
Diff-att        & -3.09  & -4.27  & -4.61  & -2.85  & -1.32  & -4.72  \\ \hline
No-rpt         & -1.51  & -2.45  & -2.18  & -1.48  & -0.44  & 0.59   \\ \hline
\end{tabular}
\caption{Ablation study results by excluding specific features from \emph{Self-att2} model. \emph{No-hyp} refers to no hypothesis, \emph{ASR} refers to no hypothesis but keeping ASR confidence, \emph{No-sent} refers to no sentence, \emph{Diff-att} denotes using different self-attention weights for the sentence and the other features, and \emph{No-rpt} denotes excluding the repetition feature. Each score is relative to the corresponding \emph{Self-att2} score in Table \ref{tab:result}.
}
\label{tab:abl_remove}
\end{table}

% \begin{table}[]
% \small
% \begin{tabular}{|c|c|c|c|c|c|c|}
% \hline
% Model  & All            & ASR           & Repeat         & IC             & HC            & SNR            \\ \hline \hline
% no hyp & 34.55          & 72.6          & 45.3           & 35.8           & 49.52         & 22.84          \\ \hline
% no utt & 51.68          & \textbf{74.2} & 56.68          & 55.2           & 67.05         & 44.99          \\ \hline
% full   & \textbf{60.76} & 73.96         & \textbf{64.71} & \textbf{58.52} & \textbf{73.2} & \textbf{56.83} \\ \hline
% \end{tabular}
% \caption{Ablation study on excluding hypothesis and sentence signal from final representation.}
%     \label{tab:abl_remove}
% \end{table}

We further explore the impact of the input features and architecture settings from our best model, \emph{Self-att2}. We represent each hypothesis as the concatenation of two vectors: the hypothesis vector and the sentence vector. Among the features of the hypothesis vector, ASR confidence is expected to be closely related to ASR and SNR ambiguities. Therefore, we conduct the following experiments: excluding whole hypothesis features (\emph{i.e.}, all the input features except the sentences), excluding hypothesis features but ASR confidence, and excluding sentence vectors.
Another architecture decision is using single self-attention weight for the concatenation of the hypothesis vector and the sentence vector. Since those two vectors are significantly different views of a hypothesis, we check if using single attention weight is better than having separate attention weights for the two different vectors.
In addition, we check the effectiveness of using the repetition feature since we hypothesize that whether the current utterance is repeated or not is helpful for the model decision.

The ablation study results are shown in Table \ref{tab:abl_remove}.
Excluding the hypothesis vector features (\emph{No-hyp}) results in a big drop in the overall performance. Hence, the hypothesis features are critical in the model decision.
Using only ASR confidence signal and the sentence vector (\emph{ASR}) is shown slightly helpful for most ambiguities except IC, but the improvements for ASR and SNR ambiguities are not very high. This means that the contextual features unrelated to the acoustics are also significantly influential to the decision for the acoustics related ambiguities.
Excluding the sentence vector (\emph{No-sent}) also causes lower performance but the drop is less compared excluding the hypothesis vector. This indicates that both the sentence and the hypotheses are important but the hypothesis features would provide more information for the model decision.
Using different self attention weights (\emph{Diff-att}) shows worse performance, which demonstrates that holistic self attention for the concatenation of the hypothesis vector and the sentence vector is not only simpler but also empirically more effective. One possible reason of the result is that the attention over different sentence vectors would be less influential because sentence vectors in different hypotheses are different only when ASR ambiguity exists.
Excluding the repetition feature (\emph{No-rpt}) also shows significant degradation, which reflects its effectiveness.

These findings show that the utilized signals and the architecture decisions are helpful for making proper model predictions.

\section{Conclusion}
\label{sec:conclusion}
 In this work, we have introduced five common ambiguities in SLU, where empirically 23\% of the utterances with these ambiguities need be clarified. To decide whether asking a clarifying question would be helpful, we have proposed a scalable neural self-attentive model, where the top and the alternative hypotheses, ambiguity occurrence information, and the other contextual information are used as the input representation and then the model predicts whether to ask clarifying questions or not. The model is supervised leveraging a user satisfaction model in order to ask a clarifying question when it would be helpful. The proposed model utilizing self-attention for hypothesis representations and ambiguity related contextual information has showed significantly improved performance compared to various baseline approaches evaluated on the user log data from a conversational agent system.

As future work, we will study how logged clarifying interactions can be utilized for the fine-tuning to further improve user satisfaction with clarification as briefly described in Section~\ref{ssec:gt_labeling}. Also, we will look at how the clarifying questions should be composed for each ambiguity type for effective and natural user engagement in the large-scale setting.

%\vfill
%\pagebreak

\bibliographystyle{IEEEbib}
\bibliography{main}

\end{document}